# 基于短语窗口的标注规则和识别算法研究

刘广[1], 涂刚[1], 李政[1], 刘译键[1]

（1. 华中科技大学 计算机科学与技术学院,湖北 武汉 430074）

通讯作者：涂刚, E-mail: tugang@hust.edu.cn

**摘要**：目前，自然语言处理大多是借助分词结果进行语法依存分析，采用的主要是基于监督学习的深度端到端方法。这种方法存在两个主要问题：首先是标注规则复杂，标注数据困难，而且工作量大；其次是算法不能识别语言成分的多粒度与多样性。为了解决这两个问题，提出了基于短语窗口的标注规则，同时设计了相应的短语识别算法。该标注规则以短语为最小单位，把句子分成 7 类可嵌套的短语类型，同时标示出短语之间的语法依存关系。对应的算法，借鉴了图像领域识别目标区域的思想，可以发现句子中各种短语的起始与结束位置，实现了对嵌套短语及语法依存关系的同步识别。实验的结果表明，该标注规则方便易用，没有二义性；算法比端到端算法更加符合语法的多粒度与多样性特征，在 CPWD 数据集上实验，比端到端方法准确性提高约 1 个点。相应的方法应用到 CCL2018 比赛中，在中文隐喻情感分析任务中获得第一名的成绩。

**关键词**：自然语言处理；标注体系；短语识别；依存分析

**中图分类号**：TP391　　　　**文献标识码**：A

## Research on Annotation Rules and Recognition Algorithm Based on Phrase Window

LIU Guang[1], TU Gang[1], LI Zheng[1] and LIU Yi-Jian[1]

(1. School of Computer Science and Technology, Nantong University, Wuhan Hubei 430074, China)

**Abstract:** At present, most Natural Language Processing technology is based on the results of Word Segmentation for Dependency Parsing, which mainly uses an end-to-end method based on supervised learning. There are two main problems with this method: firstly, the labeling rules are complex and the data is too difficult to label, the workload of which is large; secondly, the algorithm cannot recognize the multi-granularity and diversity of language components. In order to solve these two problems, we propose labeling rules based on phrase windows, and designed corresponding phrase recognition algorithms. The labeling rule uses phrases as the minimum unit, divides sentences into 7 types of nestable phrase types, and marks the grammatical dependencies between phrases. The corresponding algorithm, drawing on the idea of identifying the target area in the image field, can find the start and end positions of various phrases in the sentence, and realize the synchronous recognition of nested phrases and grammatical dependencies. The results of the experiment shows that the labeling rule is convenient and easy to use, and there is no ambiguity; the algorithm is more grammatically multi-granular and diverse than the end-to-end algorithm. Experiments on the CPWD dataset improve the accuracy of the end-to-end method by about 1 point. The corresponding method was applied to the CCL2018 competition, and the first place in the Chinese Metaphor Sentiment Analysis Task.

## 0 引言

随着即时通信、微博、论坛、朋友圈等的快速流行,人们在网络上发布的文字信息也越来越多。这些文字信息蕴含极大的价值,它们是分析社会整体及公众个体的观点、喜好、情绪、趋势等的入口。快速准确地分析文字信息,是自然语言处理研究的目标。

目前,自然语言处理研究领域存在两个问题:首先是标注规则复杂。比如哈工大等依存类别多达几十种,如表 1 所示;这种标注规则,语言学家可以轻松掌握,但是一般的标注人员,非常难以掌握。即便掌握了,在标注数据的过程中,由于语言的多样性以及熟练程度等原因,会出现各种各样问题。这就造成了标注数据慢而且错误较多,使得监督学习的模型预测准确性难以提高;其次是算法不能识别语言成分的多样性。比如:"敌人的进攻"是个名词,但是"进攻"却是动词,传统深度端到端模型只会预测出一个类别,无法预测这种嵌套的多粒度与多类别。

表 1 依存标注类别举例

| 施事关系,<br>当事关系 | 比较角色,<br>属事角色 | 并列关系,<br>选择关系 | 连词标记,<br>的字标记 |
|---|---|---|---|
| ⋯ | ⋯ | ⋯ | ⋯ |
| 成事关系,<br>源事关系 | 结局角色,<br>方式角色 | 转折关系,<br>原因关系 | 程度标记,<br>根节点 |
| ⋯ | ⋯ | ⋯ | ⋯ |

为了解决这两个问题,本文提出了基于短语窗口的标注规则,同时设计了相应的短语识别算法。该标注规则以短语为最小单位,把句子分成动词短语、名词短语等 7 类可嵌套的短语类型,同时标示出短语之间的语法依存关系。我们使用该规则,标注了各种类型的句子数据,把这个数据集称为中文短语窗口数据集(Chinese Phrase Window Dataset,CPWD)。对应的短语识别算法,借鉴了图像领域识别目标区域的 Faster RCNN 算法思想,可发现句子中各种短语的起始与结束位置,实现对嵌套短语及语法依存关系的同步识别,对应模型称为语法窗口模型(Syntax Window Model,SWM)。实验的结果表明,该标注规则方便易用,没有二义性;SWM 模型比端到端模型更加适用于语法的多粒度与多样性特征,准确性有明显提高。

## 1 相关工作

语块分析体系最早是由 Abney 在 1991 年提出的语块描述体系[1],之后 Kudoh 等[2]提出了一种基于支持向量机的语块自动分析方法;同时,Shen 等[3]提出了一种投票分类策略,将多种不同的数据表示和多种训练模型结合在一起,根据投票分类策略确定最终结果;此外,Mancev 等[4]提出了一种处理支持向量机非凸结构的斜率损失的最小化问题的序列双向方法。在汉语的语块分析方面,周强等[5,6]构造了基于规则的汉语基本块分析器,并设计了相应的基本块规则,给出了一整套解决方案,提高了基于规则的基本块分析器的性能;此外,李超等[7]应用最大熵模型和马尔科夫模型构建了一套汉语基本块的分布识别系统。

深度学习方法出现后,短语识别研究迎来了快速发展。Chiu 等[8]使用双向 LSTM 提取文本全局特征,同时,使用 CNN 提取单词的特征,进行名词短语实体的识别;Kuru 等[9]使用 Stacked Bidirectional LSTMs 提取文本全局特征进行名词短语实体识别,取得了较大进展;侯潇琪等[10]利用深度模型,将词的分布表征作为模型的输入特征维度,用于基本短语识别任务中,比使用传统的词特征表示方法提高明显;李国臣等[11]以字作

为标注单元和输入特征,基于深层模型研究短语的识别问题,并将基于 C&W 和 Word2Vec 两种方法训练得到的字分布表征作为模型的特征参数,避免了对分词及词性标注结果的依赖;徐菁等[12]利用知识图谱,提出基于主题模型和语义分析的无监督的名词短语实体指称识别方法,同时具备短语边界检测和短语分类功能;程钟慧等[13]提出了一种基于强化学习的协同训练框架,在少量标注数据的情况下,无须人工参与,利用大量无标注数据自动提升模型性能,从非结构化大数据集中抽取有意义的名词短语。

语法依存最早是著名的法国语言学家特思尼耶尔提出,我国学者徐烈炯等[14]认为,语义角色是一个"句法—语义"接口概念,而不是单纯的语义概念;刘宇红[15]提出语义和语法双向互动的观点;孙道功[16]基于词汇义征和范畴义征的分析,研究了词汇与句法的衔接机制;亢世勇等[17]通过构建"现代汉语句法语义信息语料库",研究了义类不同的体词在施事(主语、宾语、状语)和受事(主语、宾语、状语)六个语块的分布特点。这其中还包括哈工大、腾讯、百度、清华等团队的语法分类贡献。

在语法分析方面,McDonald 等[18]提出了基于图模型的依存句法分析器 MSTParser;Nivre 等[19]提出了基于转移模型的依存句法分析器 MaltParser;Ren 等[20]对 MaltParser 依存句法分析器的 Nivre 算法进行了优化,有效的改进了在汉语中难以解决的长距离依存等问题;车万翔等[21]对 MSTParser 依存句法分析器进行了改进,使用了图模型中的高阶特征,提高了依存句法分析的精度;Chris 等[22]在基于转移模型的依存句法分析框架上运用长短时记忆神经网络,将传统的栈、队列、转移动作序列看作 3 个 LSTM 细胞单元,将所有转移的历史均记录在 LSTM 中,改进了长距离依存问题;Tao Ji 等[23]开发了一种依赖树节点表示形式,可以捕获高阶信息,通过使用图神经网络(GNN),解析器可以在 PTB 上实现最佳的 UAS 和 LAS;Yuxuan Wang 等[24]提出了一种基于神经过渡的解析器,通过使用基于列表的弧跃迁过渡算法的一种变体,进行依赖图解析,获得了较好的效果;Fried 等[25]通过强化学习来训练基于过渡的解析器,提出了将策略梯度训练应用于几个解析器的实验,包括基于 RNN 的解析器。

在语义分析方面,丁伟伟等[26]利用 CRF 在英文语料上能够利用论元之间的相互关系、提高标注准确率的特点,将其运用到中文命题库,使用 CRF 对中文语义组块分类,取得好的效果;王丽杰等[27]提出了基于图的自动汉语语义分析方法,使用哈工大构建的汉语语义依存树库完成了依存弧和语义关系的分析;王倩等[28]基于谓词和句义类型块,使用支持向量机的语义角色对句子的句义类型进行识别,也有一定的启发意义。

综上,各种方法存在两个主要问题:首先是标注规则复杂。比如哈工大等依存关系多达几十种。这就造成了标注数据慢而且错误较多,做监督学习的时候,模型预测准确性难以提高;其次是算法不能识别语言成分的多样性。基于深度端到端的模型,无法对语言的多样性进行预测,无法预测嵌套多类别。

本文提出了基于短语窗口的标注规则,发布了短语窗口数据集 CPWD,同时设计了相应的短语识别算法。该标注规则以短语为最小单位,把句子分成动词短语、名词短语等 7 类可嵌套的短语类型,同时标示出短语之间的依存关系。对应的算法,借鉴了图像领域识别目标区域的 Faster RCNN 算法思想,可以发现句子中各种短语的起始与结束位置,实现对嵌套短语及语法依存关系的同步识别,对应模型称为语法窗口模型 SWM。

## 2 短语标注规范

为了实现句子的短语识别与语法依存分析，制定了一套完整的短语标注规范。该短语标注规范不仅可以对多粒度、嵌套短语进行标注，而且可以反映短语之间的依存关系。标注规则相对简单，容易学习，并大面积推广。

### 2.1 句子语法依存关系

标注规范将句子中的短语分成：名词短语、动词短语、数量词短语、介词短语、连词短语、语气词、从句，总共 7 类基本类型。句子由短语组成， 7 类基本短语类型通过树状结构组成句子，即语法依存关系。

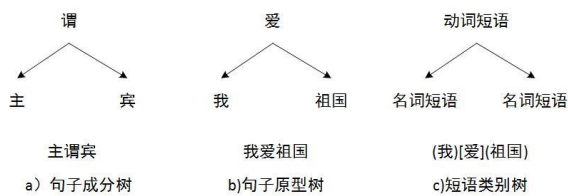

图 1 句子语法树结构图

通常，句子的树状结构由主、谓、宾关系组成，图 1 是句子语法树结构图。a)句子成分树：句子"我爱祖国"，按照句子语法可以分为主语"我"、谓语"爱"、宾语"祖国"；b)句子原型树：把"我"、"爱"、"祖国"放到对应的主谓宾位置；c)短语类别树："我"是名词短语，"爱"是动词短语，"祖国"是名词短语。

对于复杂的句子同样可以采用这种方法进行短语识别和语法依存的分析。图 2 是复杂句子的语义单元划分过程。为了方便介绍，我们使用"()"表示名词短语，"[]"表示动词短语，"{}"表示数量词短语，"<>"表示介词短语，"##"表示连词短语，"@@"表示语气词短语，"∧"表示从句。

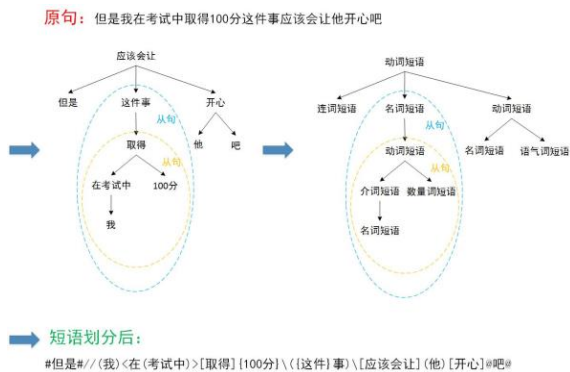

图 2 语义单元划分过程

### 2.2 短语标注规范

短语类别有名词短语"()"、动词短语"[]"、数量词短语"{}"、介词短语"<>"、连词短语"##"、语气词短语"@@"、从句"∧"。标注允许嵌套结构的存在，每种短语类别的标注规则如下。

(1) 连词短语：

连词短语是用来连接词与词、词组与词组或句子与句子、表示某种逻辑关系的虚词。连词短语可以表并列、承接、转折、因果、选择、假设、比较、让步等关系。

如："但是"表转折，"因为""所以"表因果等。

在标注体系中连词短语一般无嵌套关系。

(2) 语气词短语：

语气词短语是表示语气的虚词，常用在句尾或句中停顿处表示种种语气。

如："吗"、"吧"、"呢"等在词语、句子末，表示语气。

在标注体系中语气词短语一般无嵌套关系。

(3) 名词短语：

表示人或事物以及时间、方位等，在句子中主要充当主语、宾语、定语。它包括：以名词为中心词的偏正短语（如："伟大祖国"，"这些孩子"）；用名词构成的联合短语（如："工人农民"）；复指短语（如："首都北京"）；方位短语（如："桌面上"，"大楼前面"）；"的"字短语（如："打更

的老汉")等。某些名词短语的中心词也可以是动词、形容词，定语可以是代词、名词或其他名词短语。

(4) 动词短语：

动词短语代表动作，包括起修饰作用的状语与补语。

如："马上开始了"包括状语"马上"，中心动词"开始"，补语"了"。

(5) 介词短语：

又称为介宾短语，是介词和其它代词或名词或从句搭配形成的短语。

- "在这次考试中"为介词+名词短语，标注为<在({这次}考试中)>。
- "被"、"把"字句。如：<被(他)>，<把(他)>。

(6) 数量词短语：

数量短语，指由数词和量词组合构成的短语。

- 数量词和名词搭配，如：({一首}动听的曲子)
- 作为状语，如：[{一蹦一跳}地走着]
- 作为补语，如：[看了](他){一眼}

(7) 从句：

为了标注一个完整的语义单元，需要使用从句结构体现短语间的层次关系。

- 兼语句标为从句。如："我命令他去外面"，这里"他"既是前面的宾语，又是后面的主语，标注：(我)[命令]/(他)[去](外面)\。
- 连动句标为从句。如："我出去骑车打球"，"骑车"、"打球"为连动，标注为：(我)//[出去]\/[骑](车)\/[打](球)\\。
- 主语从句、宾语从句。如：(他)[说]/(计算机)[正在改变](世界)\。

根据该规则，我们标注了中文短语窗口数据集 CPWD，数据集包括 45,000 条从对话、新闻、法律、政策、小说中挑选的非文言文的中文句子。为了方便模型设计，句子最大长度限制在 50 个字以内。

## 3 算法

短语标注方法中，一些标签会嵌套，所以不能采用传统的端到端深度模型。为了适配新的短语标注方法，借鉴了图像领域 Faster RCNN 的算法思想，即找窗口的方法。不同的是，Faster RCNN 找目标区域窗口，本文的算法是找短语的位置窗口；Faster RCNN 会找对应的图像锚点 anchor 和 4 个偏置值，我们的算法找的是短语的锚点和两个偏置值。我们称这个算法对应的模型叫语法窗口模型（Syntax Window Model，SWM）。

### 3.1 算法流程

(1) 数据标注。例如：原句为"我爱祖国"，标注后为"(我)[爱](祖国)"。

(2) 模型输入。使用"我爱祖国"字向量作为输入。

(3) 短语特征提取。通过特征提取网络进行特征提取，可以使用传统模型进行特征提取。

(4) 短语窗口定位。特征提取结果，通过分类网络实现短语窗口定位。例如：窗口有"我"、"爱"，"祖国"三个，分类网络的输出是窗口在原句中的开始与结束位置，比如：窗口"我"的输出是(1,1)，代表在句子中的开始位置是 1，结束位置是 1。

(5) 短语分类。然后提取窗口对应的短语，通过分类网络识别出短语类别。例如："我"是名词短语，"爱"动词短语，"祖

国"名词短语。

(6) 得到结果。将短语窗口定位结果与对应的短语分类结果综合，使用标注符号表示出来，得到结果。例如：(我)[爱](祖国)。

### 3.2 SWM 模型介绍

SWM 模型主要包括字向量输入层、特征提取层、短语窗口定位层、短语窗口提取层、分类网络组成。图 3 是短语识别算法模型的基本结构图。

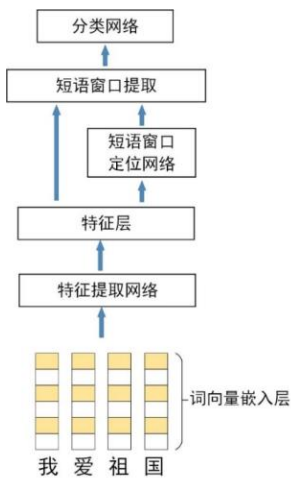

图 3 SWM 模型基本结构

模型主要包括四个组成部分：①特征提取网络：通过特征提取层实现特征的抽取，提取出的特征将被用于后续短语窗口定位层和短语窗口提取。特征提取可以使用 LSTM 或者 BERT 模型。②短语窗口定位层：短语窗口定位层用于定位短语窗口。③短语窗口提取层：收集输入的短语窗口位置 proposal、偏置值以及特征提取结果。④分类网络：使用输入的短语窗口位置，从句子中抽取对应的短语片段，对短语片段进行分类，同时会再次修正偏置值。

(1) 短语窗口定位层

短语窗口定位包括窗口穷举层、全连接层、判别和偏置层。

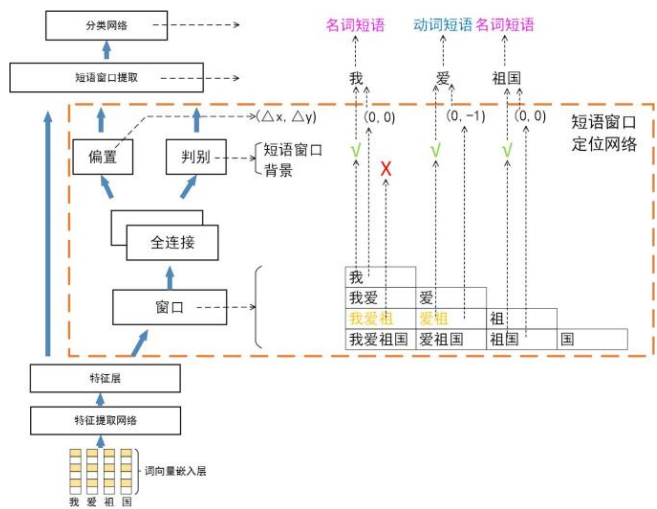

图 4 短语窗口定位网络示意图

如图 4。该层类似 Faster RCNN 的定位锚点 anchor，及推荐 proposal 的功能，可以从背景中定位锚点，也就是短语窗口，同时结合偏置值（△x 与△y）修正短语窗口的位置，得到短语窗口的 proposal。

a) 窗口穷举层

窗口穷举层用来生成候选窗口框，每个窗口由起始位置 x 和结束位置 y 组成的二元组决定，用(x,y)表示。图 5 是窗口穷举示意图，其中"我"是第一个字，它的窗口起始位置是 1，结束位置是 1，表示成(1,1)，同理，"我爱"对应(1,2)。

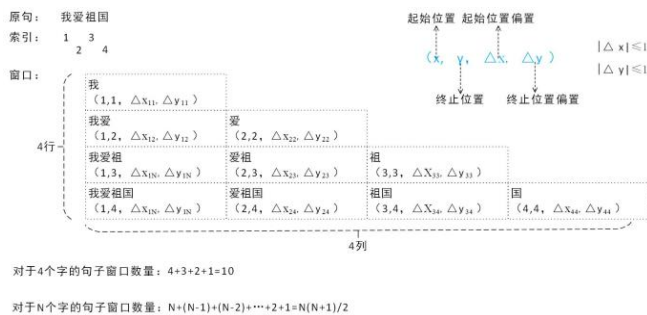

图 5 窗口示意图

采用所有的子窗口锚点 anchor 来生成候选窗口 proposal。句子总长度为 N 时，第一个字作为起始位置的窗口有 N 个，第二个字作为起始位置的窗口有 N-1 个···最后一个字作为起始位置对应的窗口数量有 1 个，所以总的窗口数量是 N(N+1)/2，对应同样数量的锚点 anchor。

b) 背景和偏置层

所有 N (N+1)/2 个子窗口的特征，经过全连接层后，输出是 N (N+1)个分类及对应的 N (N+1)个偏置值，如图 6 所示。也就是说，每个子窗口特征，经过全连接层后，输出 2 个分类的 one-hot 值，用于确定哪些短语窗口 anchor 是短语 proposal，以及 x 与 y 的偏置值△x 与△y 两个值，每个子窗口特征得到 4 个输出值。

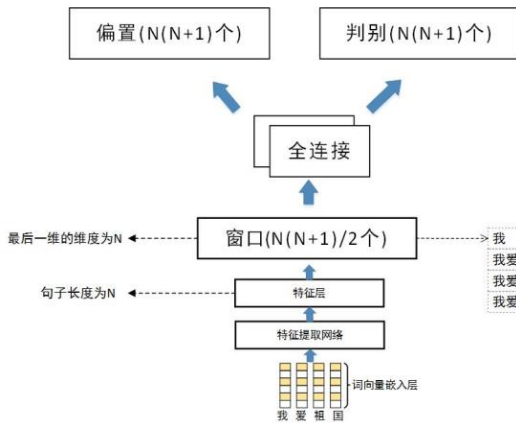

图 6 窗口判别与偏置

为了定位的窗口位置更加精确，增加两个偏置值△x 与△y，相当于一个回归准确性的过程。图 7 是偏置示意图，红色窗口"爱祖"通过全连接层后，被识别为短语，而正确的短语是绿色窗口"爱"，这就需要通过偏置值修正。把原来的窗口坐标加上偏置值，原来的窗口(x,y)变为(x+△x, y+△y)。其中△x 表示的是窗口起始位置偏置值，△y 表示的是窗口结束位置偏置值。

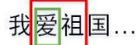

图 7 偏置示意图

偏置值计算举例：

- x=6，△x=1，修正结果：x=7
- x=6，△x=0，修正结果：x=6
- x=6，△x=-1，修正结果：x=5
- …

为了防止预测窗口 proposal 过多，规定当预测窗口坐标和真实短语窗口坐标偏差小于 1 时，窗口的偏置才是有效的。图 8 是有效偏置示意图。短语起始位置可以向左或右最多偏移一个字，终止位置也可以向左或右最多偏移一个字。当窗口长度大于等于 2 时，规定最多有 5 种偏移情况（proposal），当窗口长度等于 1 时，最多只有 3 种偏移情况（proposal）。

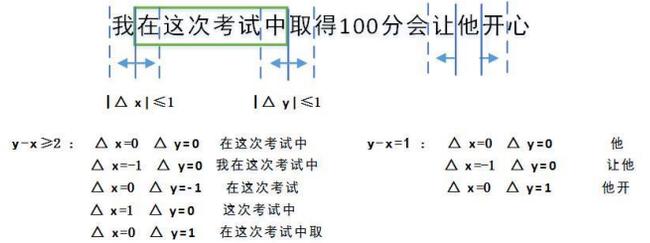

图 8 有效偏置示意图

c) 损失函数

模型对应的损失函数，包括两个。其一是分类（one-hot 向量）的损失函数，即判断是否为短语窗口的判别损失函数。采用的交叉熵损失函数。式 1 是判别损失函数，其中 M 是短语类别数量；其二是回归偏置值△x 与△y 的损失函数，即短语窗口和真实窗口之间的误差，由于是多个值，所以采用的是均方根误差 RMSE。式 2 是偏置损失函数。

$$loss_{判别} = -\sum_{c=1}^{M} y_c \log(p_c) \quad (1)$$

$$loss_{偏置} = \sqrt[2]{\frac{1}{M}\sum_{c=1}^{M}(y_c - \hat{y}_c)^2} \quad (2)$$

(2) 分类网络

分类网络利用已经获得的短语窗口 proposal 和偏置，通过全连接层与 softmax 层推算每个短语窗口具体属于哪个类别（如：介词短语，名词短语，数量词短语等），输出每个短语类别的概率向量，得到单一分类标签；同时再次利用回归偏置获得每个短语的位置偏移量，用于微调的短语窗口位置。

# 4 实验

## 4.1 数据集与评估标准

实验数据集使用标注的中文短语窗口数据集 CPWD，包括 45,000 条从对话、新闻、法律、政策、小说中挑选的中文句子。其中文言文只占不到 5%比例，多是一些成语与谚语组成的句子。为了方便模型设计，句子最大长度限制在 50 个字以内。

传统端到端的结果统计，主要根据每个字预测得到标签的情况。这样的统计方式存在偏差，不如按照短语统计准确，比如："中华人民共和国"命名实体的标签是"BIIIIII"为正确，如果预测结果是"BIIIBII"，那么存在一个标签错误。按照传统方式统计，只算 7 个标签中出现了一次错误；按照短语方式统计，"BIIIIII"全对为正确，"BIIIBII"为错误，即有一个标签错误，整个短语的预测是错误的，这样更加准确合理。

## 4.2 实验环境

实验采用 Python 语言实现，python 版本为 3.6.1。使用的框架为 TensorFlow，版本为 1.12.0。使用的电脑配置为内存：32G，处理器：Intel Xeon(R) CPU E5-2623 v3 @3.00GHz*8，显卡：TITAN Xp，操作系统类型：ubuntu14.04 64-bit。

## 4.3 实验分析

首先，SWM 在多种网络结构的情况下进行对比优化。优化方法包括采用双层 BiLSTM，采用 BERT 代替 BiLSTM，使用 CRF 层，在 BiLSTM 层之前加入 CNN 层进行特征抽取，选取不同比例的反例。结果如表 2 所示。

表 2 SWM 优化过程对比

|  | Acc | F1 | 前向传播时间 |
| --- | --- | --- | --- |
| 1 层 BiLSTM | 88.35 | 86.68 | 65 |
| 2 层 BiLSTM | 88.24 | 87.13 | 80 |
| BERT | 88.79 | 88.65 | 106 |
| CNN+BiLSTM | 87.42 | 87.39 | 96 |
| 1:1 正反例 | 87.01 | 87.63 | 87 |
| 1:2 正反例 | 88.17 | 88.37 | 89 |

从结果可以看出，最优的模型结构是：1:2 正反例，BERT+CRF。但是考虑到运行效率和资源占用情况，我们在做实验或者工程部署的时候，建议采用的特征层模型是：1:2 正反例，2 层 BiLSTM。

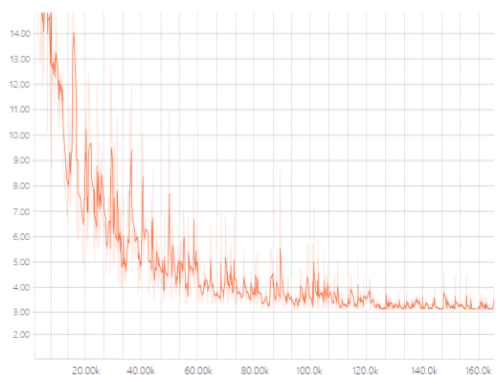

（a）BERT

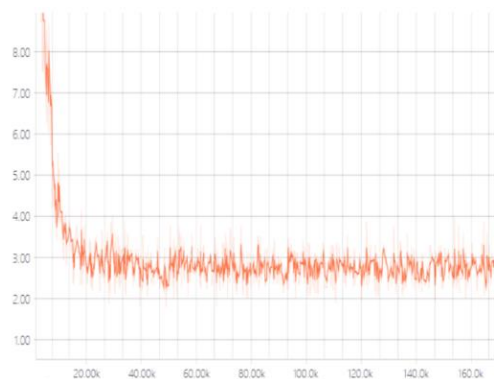

（b）BiLSTM

图 9 Loss 收敛情况图

Loss 值收敛情况在 BERT、BiLSTM 两种最优模型之间进行比较，收敛情况如图 9。可以看到，BiLSTM 收敛快些，BERT 语言模型收敛慢些，而且一个 epoch 的训练时间也长一些。

## 4.4 对比实验结果

SWM 对比各种端到端算法。

由于各种端到端算法输出与标签形式有不同，所以将标签形式调整成在统一的方式下进行对比。SWM 模型输出标签比端到端多，附带有嵌套等信息，所以需要进行降维处理，之后与端到端模型进行对比。降维后可以形成命名实体标签、依存标签，分别与 BiLSTM、BERT 等端到端算法进行对比。

表3 实验对比结果

|  | 命名实体 F1 | 依存分析 F1 |
| --- | --- | --- |
| BiLSTM | 88.12 | 86.70 |
| BiLSTM+CRF | 89.06 | 87.99 |
| BERT | 90.82 | 89.20 |
| BERT+CRF | 90.32 | 89.10 |
| CNN+CRF | 88.41 | 87.89 |
| SWM（BiLSTM） | 90.31 | 89.26 |
| SWM（BERT） | **91.37** | **90.85** |

可以看到 SWM 模型比传统端到端有优势。这种优势的产生，我们分析是由于 SWM 更适合语言的嵌套特征，使得模型不用在多个嵌套的命名实体之间做多选一的抉择，降低了模型的困惑度，解耦了文字与标签一对一的限制。

## 5 总结

本文针对传统自然语言处理存在的问题进行改进。包括两个方面的问题：标注方法过于复杂，对应的深度端到端算法无法解决语法多样性问题。首先，定义了基于短语窗口的标注方法，然后，标注了中文短语窗口数据集 CPWD，最后，给出了对应的语法窗口识别算法模型 SWM。新的标注规则以短语为最小单位，把句子分成 7 类可嵌套的短语类型，同时标示出短语之间的语法依存关系，易于实施。SWM 模型，借鉴了图像领域识别目标区域（Faster RCNN）的思想，可以发现句子中各种短语的起始与结束位置，实现了对短语及语法依存关系的同步识别，解决了嵌套短语问题。实验的结果表明，该标注规则方便易用，没有二义性；SWM 模型比端到端模型更加契合语法的多粒度与多样性特征，提高了准确性。

## 参考文献